# Falkor-IRAC: Graph-Constrained Generation for Verified Legal Reasoning in Indian Judicial AI

Joy Bose

*Independent Researcher, Bengaluru, India*

## Abstract

Legal reasoning is not semantic similarity search. A court judgment encodes constrained symbolic reasoning: precedent propagation, procedural state transitions, and statute-bound inference. These are properties that vector-based retrieval-augmented generation (RAG) cannot faithfully represent. Hallucinated precedents, outdated statute citations, and unsupported reasoning chains remain persistent failure modes in LLM-based legal AI, with real consequences for access to justice in high-caseload jurisdictions such as India. This paper presents Falkor-IRAC, a graph-constrained generation framework for Indian legal AI that grounds generation in structured reasoning over an IRAC (Issue, Rule, Analysis, Conclusion) knowledge graph. Judgments from the Supreme Court and High Courts of India are ingested as IRAC node structures enriched with procedural state transitions, precedent relationships, and statutory references, stored in FalkorDB for low-latency agentic traversal. At inference time, LLM-generated answers are accepted only if a valid supporting path can be traced through the graph, a check performed by a falsifiability oracle called the Verifier Agent. The system also detects doctrinal conflicts as a first-class output rather than silently resolving them. Falkor-IRAC is evaluated using graph-native metrics: citation grounding accuracy, path validity rate, hallucinated precedent rate, and conflict detection rate. These metrics are argued to be more appropriate for legal reasoning evaluation than BLEU and ROUGE. On a proof-of-concept corpus of 51 Supreme Court judgments, the Verifier Agent correctly validated citations on completed queries and correctly rejected fabricated citations. Evaluation against vector-only RAG baselines is left for future work, as is GPU-accelerated inference to address current timeout rates on CPU hardware. The companion InIRAC dataset, 500+ structured Indian court judgments with IRAC annotations, procedural event chains, and typed precedent relationships, is released alongside this paper at https://huggingface.co/datasets/joyboseroy/inIRAC.

***Keywords***: *legal AI, knowledge graphs, graph-constrained generation, hallucination reduction, Indian judiciary, IRAC, FalkorDB, verified generation, multi-agent reasoning*

## 1. Introduction

Courts are graph traversal engines disguised as prose. A Supreme Court judgment is not a document in the retrieval sense; it is a traversal record, evidence that a judge walked a path through a graph of issues, rules, precedents, and statutory provisions and arrived at a conclusion traceable back to every node along the way.

The dominant approach in legal AI treats this traversal record as a retrieval problem. Judgment text is chunked, embedded in a vector space, and retrieved by semantic similarity when a query arrives. This is Retrieval-Augmented Generation (RAG), and it has produced useful systems in many domains. For law, it fails in three structural ways.

First, semantic similarity is not legal relevance. A landmark precedent may use entirely different vocabulary from the case before the court; its relevance is structural, not lexical. Second, hallucinated precedents are a professional liability. A plausible-sounding citation that does not exist is not a quality failure; it is a defect that can directly harm the person relying on it. Third, legal reasoning is stateful. A query is embedded in a procedural process whose current state determines which rights attach, which time limits run, and which precedents govern. RAG has no model of this state.

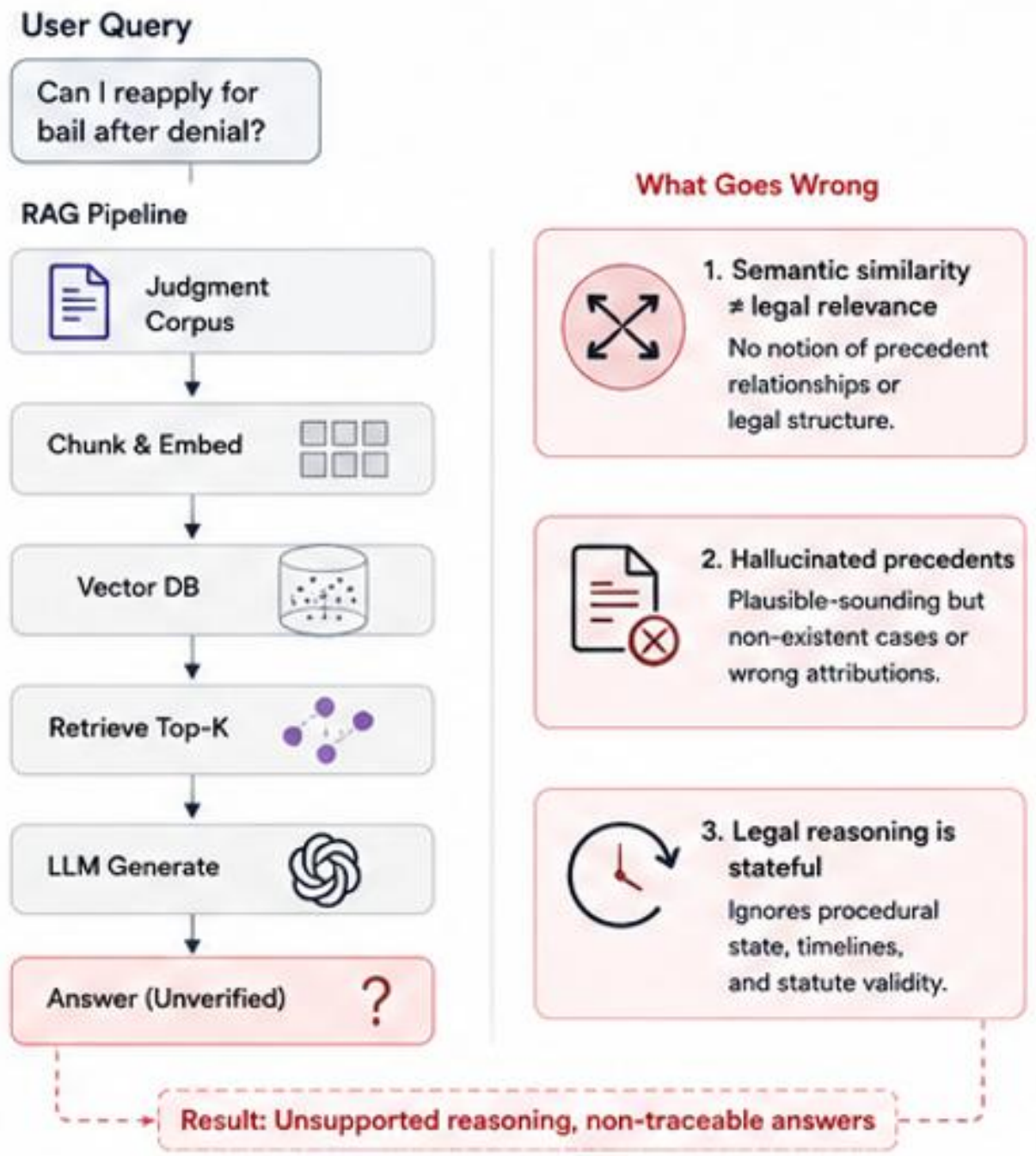

*Figure 1: Why vector RAG fails in legal reasoning. The standard pipeline (left) retrieves by semantic similarity, generates without constraint, and returns an unverified answer. Three structural failure modes are shown: semantic similarity is not legal relevance; hallucinated precedents are a professional liability; and legal reasoning is stateful, requiring knowledge of where a litigant currently sits in the procedural sequence.*

India provides a demanding test case. With approximately 50 million pending court cases, chronically underfunded legal aid, and a population that interacts with the legal system in multiple

Indic languages, the access-to-justice gap is severe. A voice-first legal AI capable of explaining court notices, tracking procedural state, and surfacing relevant precedents without hallucination would serve a genuine need that existing tools do not address.

This paper presents Falkor-IRAC, a graph-constrained generation framework that addresses these failures. The key architectural choice is the separation of generation from verification: the LLM proposes an answer; a Verifier Agent checks whether a valid citation path supporting that answer exists in a FalkorDB knowledge graph. If no such path exists, the answer is rejected. This is not retrieval-augmented generation. It is generation constrained by graph.

The contributions are as follows:

1. An IRAC knowledge graph schema for Indian judicial reasoning, extended with procedural state transitions and conflict-typed precedent relationships. This is the first such schema to model litigation flow alongside doctrinal reasoning.
2. A graph-constrained generation architecture in which LLM outputs are accepted only if a valid citation path exists in the graph, with iterative revision on failure and explicit abstention when no verified answer is available.
3. Conflict detection as a first-class output. The system surfaces coordinate bench disagreements, per incuriam citations, and fact-based distinctions rather than resolving them silently.
4. A graph-native evaluation suite covering citation grounding accuracy, path validity rate, hallucinated precedent rate, and conflict detection rate, proposed as a standard for legal reasoning benchmarks where BLEU and ROUGE are inadequate.

The paper is organised as follows. Section 2 reviews related work. Section 3 describes the Falkor-IRAC framework. Section 4 presents the experimental setup. Section 5 reports results. Section 6 discusses implications and limitations. Section 7 concludes.

## 2. Background and Related Work

### 2.1 Indian Legal NLP

The Indian Legal Documents Corpus (ILDC, Malik et al. 2021) provided the first large-scale annotated corpus of Supreme Court judgments for prediction and explanation tasks and remains the standard benchmark for Indian legal NLP. InLegalBERT, a BERT model pre-trained on Indian legal text, has over 1.8 million downloads on HuggingFace. NyayaAnumana (Law-AI Lab, IIT Kharagpur) is the largest Indian legal judgment prediction dataset. INLegalLlama extends LLaMA through continual pre-training on Indian legal documents, achieving approximately 90% F1-score on judgment prediction. IndicLegalQA (2025) provides a question-answering dataset for the Indian judicial context, and MILPaC covers English and nine Indic languages in a parallel legal corpus.

None of these systems enforce grounded verification of generated reasoning chains. They predict or summarise judgments; they do not check whether the reasoning they produce is traceable to real precedents. Falkor-IRAC addresses this gap directly.

### 2.2 Legal Knowledge Graphs

Prior knowledge graph work in legal AI has largely focused on citation networks: Case nodes connected by CITES relationships. This is structurally shallow. It records that one case cited another without capturing for what proposition, with what authority, or subject to what subsequent treatment. Work on COLIEE, ECHR datasets, and LegalBench has advanced legal NLP benchmarking but in non-Indian jurisdictions and without procedural modelling. No prior Indian legal knowledge graph includes conflict-typed relationships (CONFLICTS_WITH, RESOLVED_BY, NARROWED_BY) as first-class edges, nor procedural event chains alongside doctrinal reasoning. Falkor-IRAC is the first Indian legal knowledge graph to include both.

### 2.3 RAG and Its Limits in Legal Reasoning

Standard RAG (Lewis et al., 2020) retrieves document chunks by vector similarity and conditions generation on the retrieved context. GraphRAG (Microsoft, 2024) improves retrieval by traversing a graph of entities and relationships before generating. The distinction between GraphRAG and Falkor-IRAC is architectural: GraphRAG uses the graph to improve what the model sees at retrieval time; Falkor-IRAC uses the graph to constrain what the model is permitted to claim at generation time. The difference is between a better library and an external examiner with veto power.

### 2.4 Hallucination in Legal AI

Han (2026) surveys trustworthy legal reasoning and identifies grounding and verifiability as the two primary open problems. Song et al. (2026) propose knowledge graph-assisted LLM post-training to reduce legal hallucination, focusing on fine-tuning rather than inference-time constraint. Karna (2026) proposes a hybrid RAG-LLaMA framework for Indian legal text. Falkor-IRAC differs from all three in operating at inference time with hard graph-path constraints, requiring no additional training beyond the base LLM.

### 2.5 Verified and Constrained Generation

Constitutional AI (Anthropic, 2022) applies soft normative constraints through critique and revision. Self-RAG (Asai et al., 2023) introduces retrieval and reflection tokens enabling selective retrieval and self-critique. Both use soft constraints: the model is guided toward compliance but not hard-blocked on violation. The Falkor-IRAC Verifier uses a hard constraint: path non-existence in the graph is a veto, not a penalty. In legal contexts, a plausible but unverifiable answer is worse than no answer, which is why the hard constraint is the appropriate design.

## 3. The Falkor-IRAC Framework

### 3.0 Legal Reasoning as Constrained Graph Traversal

Before describing the system, it is useful to state the theoretical framing precisely. Legal reasoning is treated here as a problem of constrained path traversal over a directed graph, not as a problem of semantic similarity search over a corpus.

Let $G = (V, E)$ be a directed graph where $V$ is the set of legal entities (cases, statutes, sections, issues, rules, procedural events) and $E$ is the set of typed relationships between them (CITES, APPLIES_RULE, TRIGGERS, CONFLICTS_WITH, and so on).

A legal claim c is a proposition of the form: "conclusion X follows from rule R applied to facts F under precedent P." A support path for c is a sequence of nodes and edges P_c = (v_1, e_1, v_2, e_2, ..., v_n) in G such that the path grounds every element of c in a node or relationship in G.

Validity is then defined as:

```
Valid(c) = 1  iff  there exists P_c in G such that P_c |= c
```

Where |= denotes that the path P_c provides sufficient grounding for claim c under the schema constraints (correct relationship types, no OVERRULED nodes, no STALE statutes).

From this definition, two key evaluation metrics follow directly. The hallucination rate H over a set of generated claims C is:

```
H = |C_invalid| / |C_total|
```

Where C_invalid is the subset of claims for which no valid support path exists. The path validity rate (PVR) over a set of generated answers A is:

```
PVR = |A_verified| / |A_generated|
```

Where A_verified is the subset of answers for which every claimed citation has a valid support path.

This framing clarifies why hard veto is the appropriate design. Under the probabilistic view of generation, a claim without a support path is not a low-confidence claim; it is a claim with no grounding in the legal graph at all. Treating it as a soft penalty would allow ungrounded claims to appear in output with reduced confidence scores. The hard veto eliminates them.

One important scope qualification: validity as defined here is relative to the ingested graph G. A missing path means the claim cannot be grounded in G, not that it has no basis in law. The system makes no claim to have verified against all of Indian case law, only against the ingested corpus. This distinction is made explicit in the system output and discussed further in Section 7.

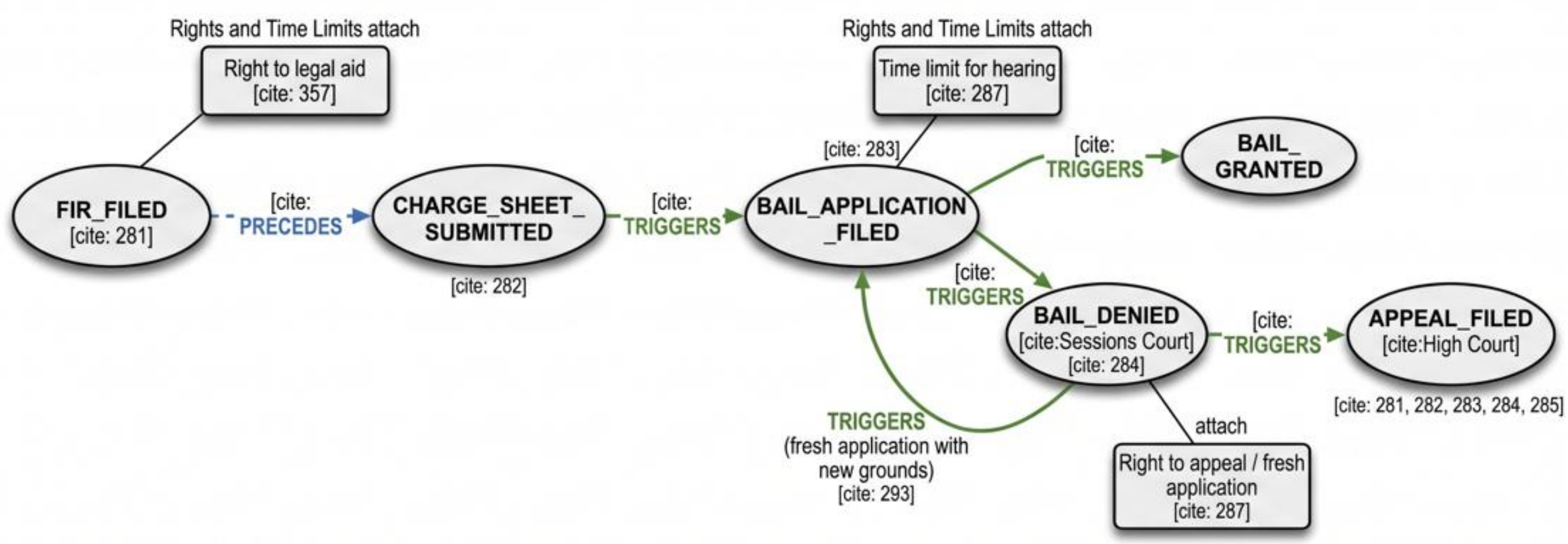

*Figure 2: Procedural State Machine for Indian Bail Matters. This diagram illustrates the "litigation flow" where nodes represent procedural states (e.g., BAIL_DENIED) and edges represent transitions (TRIGGERS, PRECEDES). Unlike vector-based retrieval, this graph-grounded model tracks a litigant's specific position in the legal process to determine which rights (e.g., right to legal aid) or time limits apply at any given transition. Note the recursive TRIGGERS relationship from BAIL_DENIED back to a fresh application, modeling the requirement for "new grounds".*

### 3.1 Knowledge Graph Schema

The knowledge graph is stored in FalkorDB, a graph database supporting hybrid vector-plus-graph queries at sub-millisecond latency. The schema is built around three layers absent from prior Indian legal knowledge graph work: IRAC structure, procedural event chains, and conflict-typed precedent relationships.

Node types are as shown in table 1.

| Node Type | Description |
|---|---|
| Case | Judgment node with citation, court, bench size and type, matter type, summary |
| Judge | Author or bench member with tenure metadata |
| Statute | Act or ordinance with repeal status |
| Section | Specific provision with repeal status and amendment links |
| LegalIssue | The question framed by the court, typed by category |
| Rule | Legal principle extracted from the judgment |
| Argument | Petitioner or respondent position |
| ProceduralEvent | Typed event in the litigation timeline (bail, appeal, delay, etc.) |
| Outcome | Court conclusion with outcome type |
| Jurisdiction | Court and territorial scope |

*Table 1: Knowledge Graph Node Types and Descriptions.*

Relationship types include standard citation relationships (CITES, OVERRULES, DISTINGUISHES) and three novel types introduced by this work. Table 2 lists the relationship types.

| Relationship | Key Attributes | Novel |
|---|---|---|
| CITES | proposition: string | |
| OVERRULES | year: integer | |
| DISTINGUISHES | basis: string | |
| CONFLICTS_WITH | conflict_type: coordinate_bench \| per_incuriam \| distinguished; unresolved: bool | Yes |
| RESOLVED_BY | resolution_type: larger_bench \| full_bench \| constitutional_bench | Yes |
| NARROWED_BY | basis: string | Yes |
| TRIGGERS | condition: string (procedural) | Yes |
| PRECEDES | time_gap_days: integer (procedural) | Yes |
| APPLIES_RULE | | |
| RESULTS_IN | | |

*Table 2: Relationship Types and Metadata*

The procedural event layer (ProceduralEvent nodes connected by TRIGGERS and PRECEDES) distinguishes this schema from citation-network approaches. It enables reasoning over litigation timelines, answering questions about procedural state such as what options are available after a bail denial by a coordinate bench. This is a question that cannot be answered from a citation graph alone.

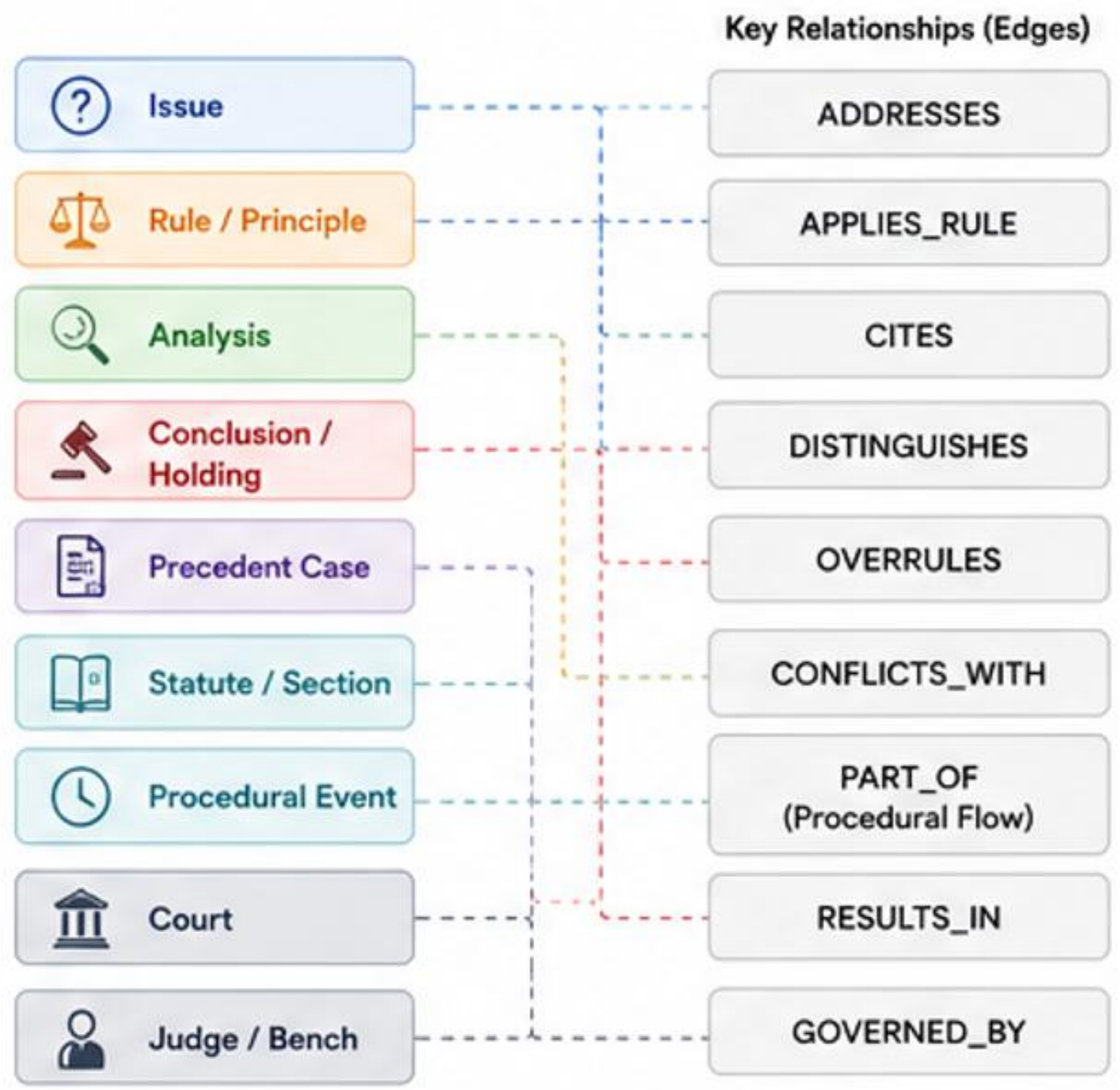

*Figure 3: IRAC knowledge graph node types and relationship types. Left column shows nine node types covering the IRAC structure (Issue, Rule, Analysis, Conclusion) and supporting entities (Precedent Case, Statute/Section, Procedural Event, Court, Judge/Bench). Right column lists the corresponding edge types: ADDRESSES, APPLIES_RULE, CITES, DISTINGUISHES, OVERRULES, CONFLICTS_WITH, RESULTS_IN, and GOVERNED_BY.*

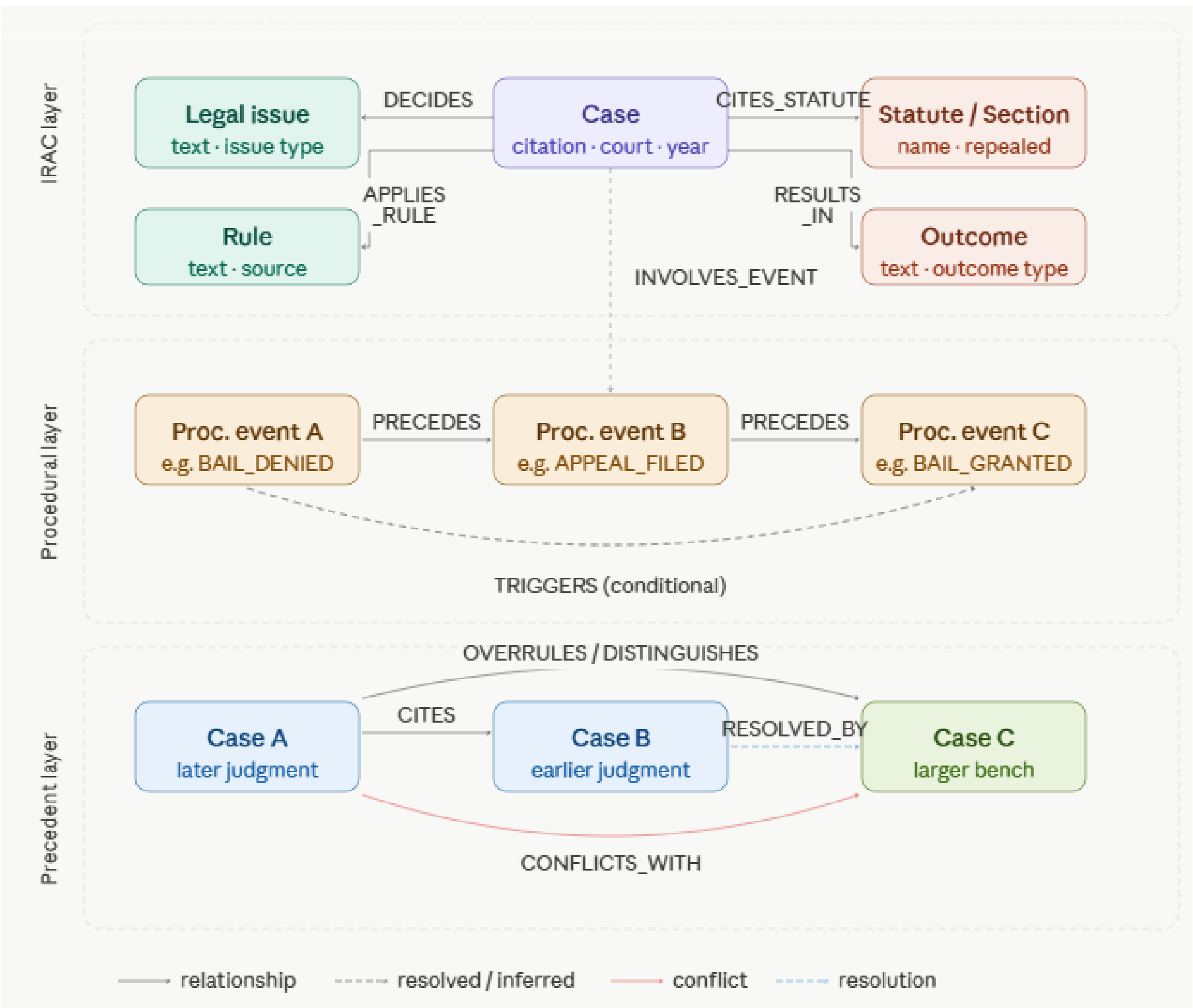

*Figure 4: Three-layer knowledge graph schema. Top layer (IRAC): the central Case node connects to LegalIssue, Rule, Statute/Section, and Outcome. Middle layer (procedural): ProceduralEvent nodes chained via PRECEDES and TRIGGERS. Bottom layer (precedent): Case nodes linked via CITES, OVERRULES, DISTINGUISHES, and the novel CONFLICTS_WITH and RESOLVED_BY edges that surface unresolved doctrinal splits.*

### 3.2 Ingestion Pipeline

Judgment PDFs are processed through a three-stage pipeline. Stage one extracts clean text using PyMuPDF with layout-aware block sorting to handle multi-column layouts and remove headers and footers. Best-effort metadata extraction detects the case citation, court, bench composition, and year from the first 2,000 characters, where Indian court judgments consistently place this information.

Stage two uses an LLM to extract IRAC structure from the cleaned text. A constrained JSON prompt asks the model to identify issues, rules, analysis summary, conclusion, cited statutes, cited precedents with relationship types, and procedural events. The model is instructed not to invent citations; fields are left empty rather than fabricated when confidence is low. Stage three loads the structured JSON into FalkorDB using idempotent MERGE operations, so that repeated ingestion of the same judgment does not create duplicate nodes.

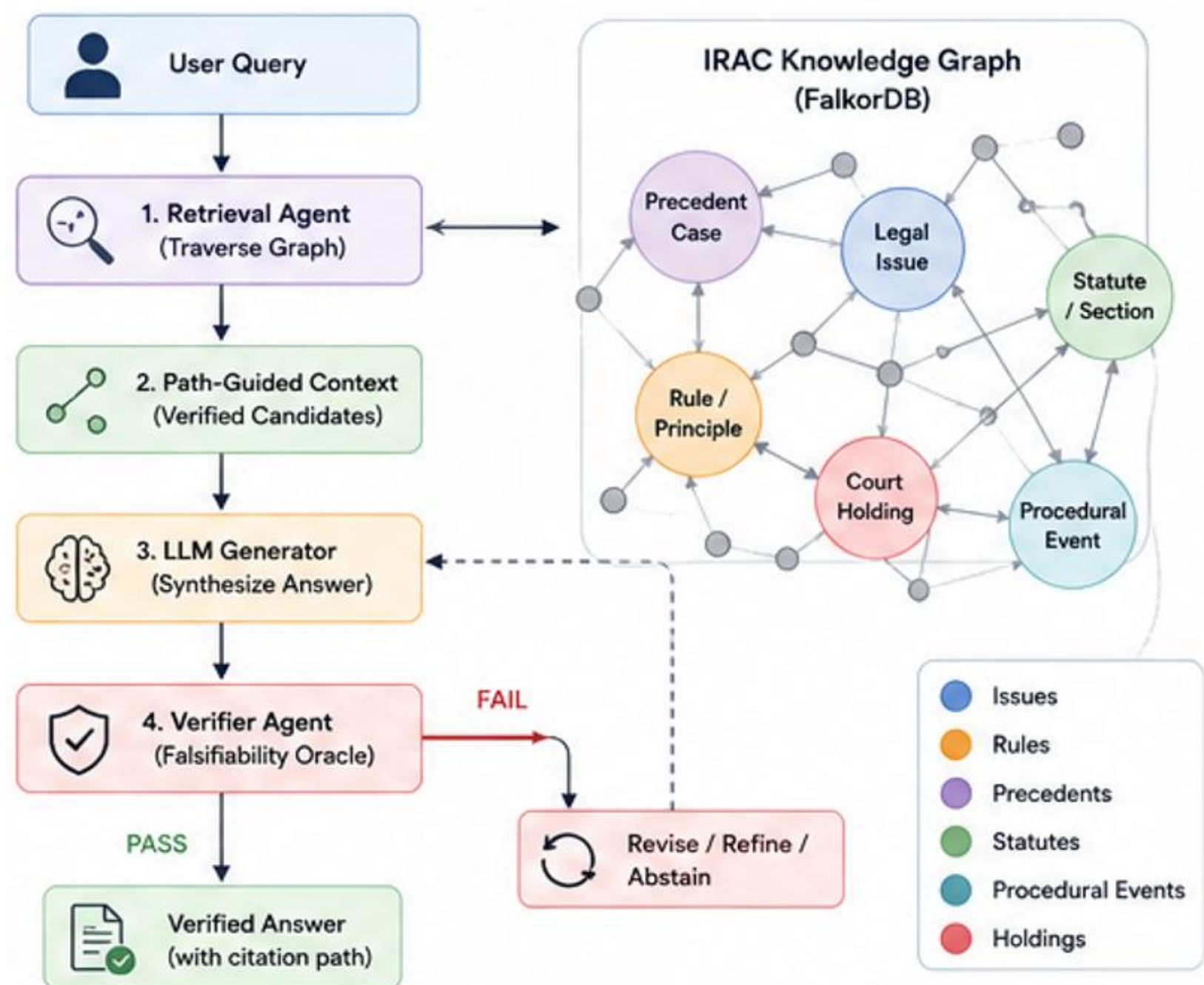

*Figure 5: The Falkor-IRAC graph-constrained generation architecture. The Retrieval Agent traverses the IRAC Knowledge Graph in FalkorDB and returns path-guided context. The LLM Generator synthesises an answer. The Verifier Agent (falsifiability oracle) checks whether a valid citation path exists in the graph. On failure the system revises or abstains; on pass the verified answer with its citation chain is returned. The graph is the constraint, not merely the retrieval index.*

### 3.3 Graph-Constrained Generation Architecture

The full inference pipeline is as follows:

```
User Query
     |
     v
Retrieval Agent  (graph traversal, precedent candidates)
     |
     v
LLM Synthesis    (path-guided answer generation)
     |
     v
Verifier Agent   <-- falsifiability oracle
     |
     +-- VALID    --> Answer + citation chain returned
     +-- INVALID  --> Revise (up to 2 attempts) or abstain
     +-- CONFLICT --> Conflict metadata returned with answer
     +-- STALE    --> Statute repeal flagged, revise or abstain
```

The Retrieval Agent traverses FalkorDB using five strategies in parallel: matter-type traversal, statute-section traversal, issue-keyword traversal, citation-chain expansion, and conflict detection among candidates. Results are ranked by court authority (Supreme Court over High Court) and recency, then deduplicated.

The LLM synthesis step is path-guided: the model is given the retrieved precedent candidates with their graph-derived metadata and instructed to cite only from the provided list. This is not a guarantee of correctness, but it is a strong prior toward grounded generation.

The Verifier Agent is the core component. Given the proposed answer and the citation chain it claims to rely on, it queries FalkorDB to check path existence. For each cited case, it verifies existence in the graph, checks for subsequent overruling, and checks for conflict relationships with other cited cases. For each cited statute, it checks repeal status. The Verifier returns one of four outputs: VALID, INVALID, CONFLICT, or STALE.

On INVALID or STALE, the LLM is prompted to revise. The revision prompt includes the rejection reason and instructs the model to cite only verified sources or to abstain if no verified answer is available. After two failed revision attempts, the system abstains rather than returning an unverified answer.

A scope note on the verification guarantee: the Verifier checks citation paths against the ingested corpus, not against all of Indian law. VALID status means the cited cases exist in the graph and no overruling or conflict has been detected within the corpus. It does not mean the answer is correct as a matter of law; it means it is grounded in the available evidence. This distinction is important for honest deployment and is reflected in the evaluation metrics and in the system output text.

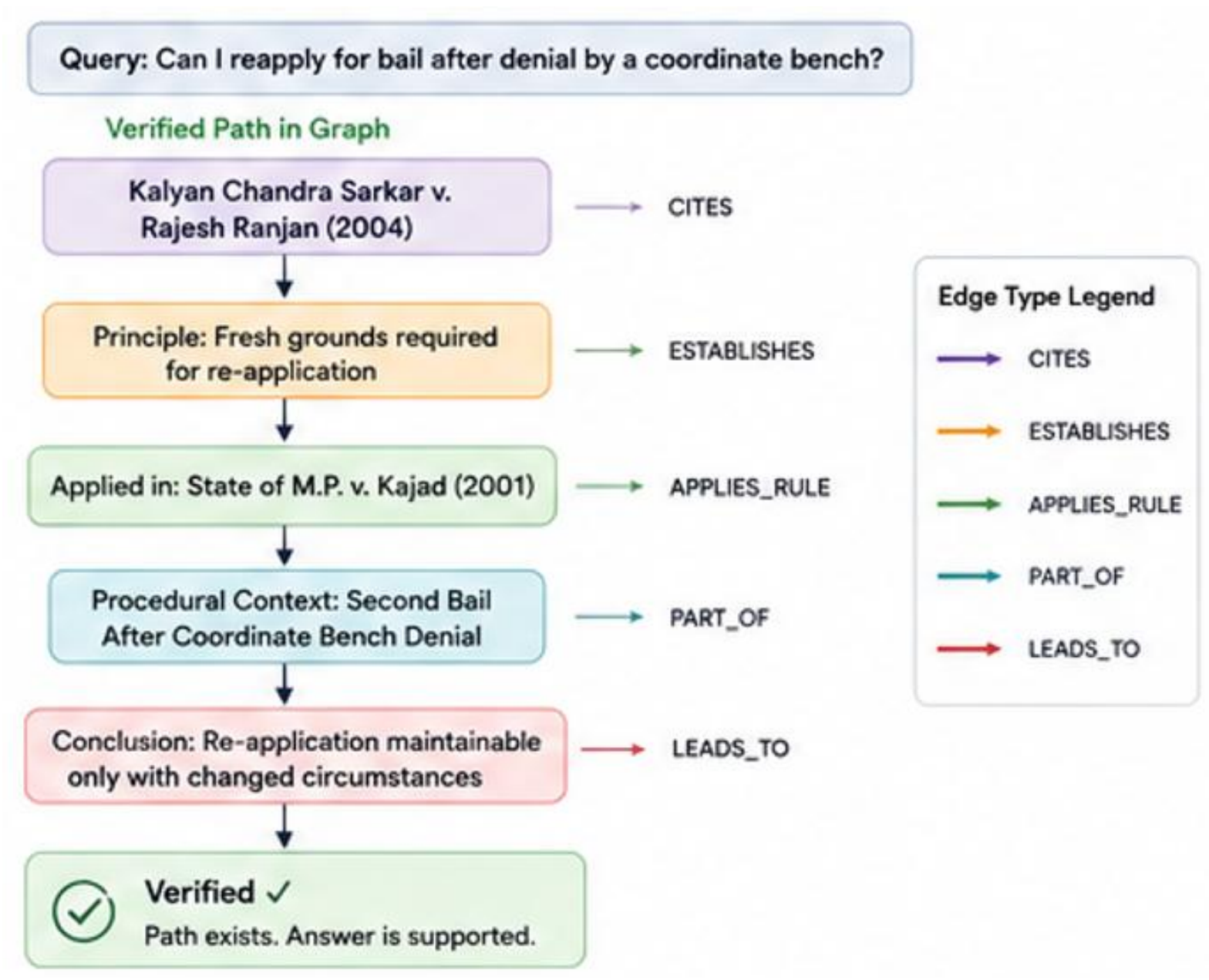

*Figure 6: Verified reasoning path for a bail reapplication query. The path traces from Kalyan Chandra Sarkar v. Rajesh Ranjan (2004) through the principle that fresh grounds are required, to its application in a later case, and to the conclusion that re-application is maintainable only with changed circumstances. Each node and edge is confirmed in the graph before the Verifier returns VALID with confidence 1.0.*

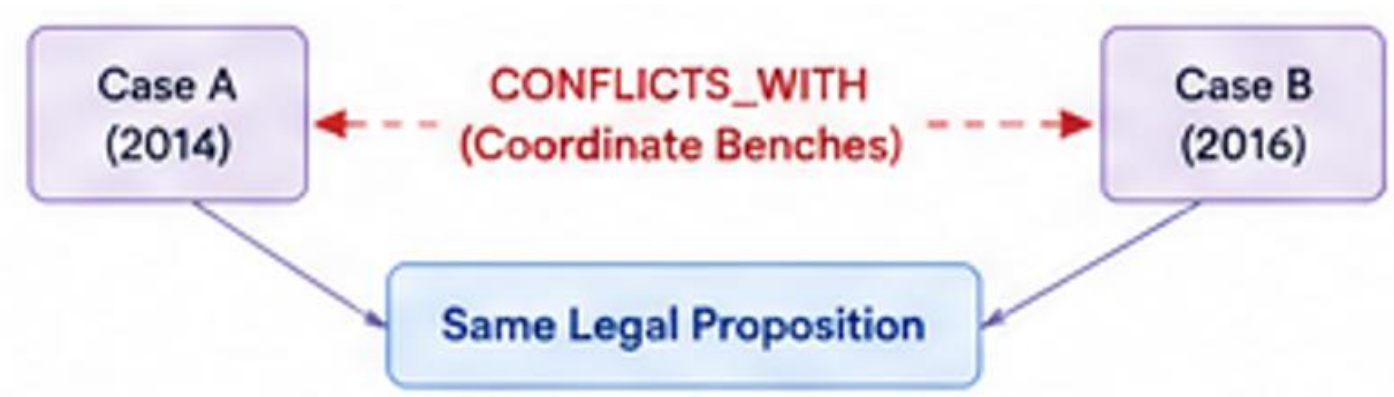

*Figure 7: Conflict detection as a first-class output. Case A (2014) and Case B (2016), both coordinate bench decisions, are connected by a CONFLICTS_WITH edge on the same legal proposition. Because no RESOLVED_BY edge exists, the Verifier returns CONFLICT status and surfaces both paths with conflict metadata rather than silently choosing one side. For a legal practitioner, explicit conflict disclosure is more useful than a confident answer drawn from an unresolved split.*

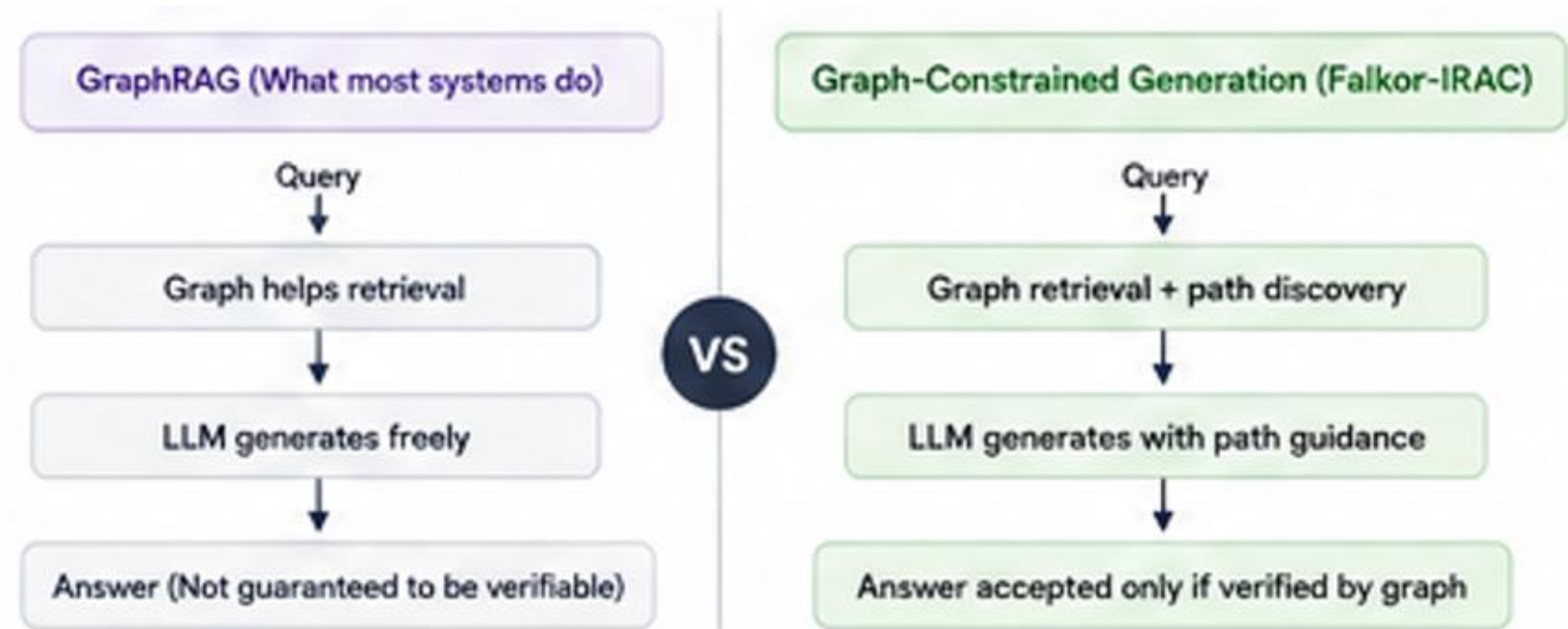

*Figure 8: GraphRAG versus graph-constrained generation. In GraphRAG (left), the graph improves retrieval but the LLM generates freely and the answer is not guaranteed to be verifiable. In Falkor-IRAC (right), the LLM generates with path guidance and the answer is accepted only if a valid supporting path exists in the graph. The distinction is between a better library and an external examiner with veto power.*

## 3.4 Conflict Detection

Doctrinal conflict in Indian courts arises in three forms: coordinate bench disagreements (two benches of the same size reach opposite conclusions on the same statutory provision), per incuriam citations (a case cites a precedent without noticing it was overruled), and fact-based distinctions applied so broadly they effectively conflict with the original holding. Falkor-IRAC models all three through typed CONFLICTS_WITH relationships.

When the Verifier detects that two or more of the proposed citations are connected by an unresolved CONFLICTS_WITH edge, it returns CONFLICT status with the following output:

```
{
```

```
  "answer": "Based on current precedent...",
  "supporting_paths": ["(2012) 9 SCC 1", "(2013) 4 SCC 20"],
  "conflict": true,
  "conflict_type": "coordinate_bench",
  "resolution": "unresolved - refer to larger bench ruling if available",
  "confidence": "low"
}
```

The system never silently resolves doctrinal conflict. For a pro se litigant, honest uncertainty is more useful than a confident answer drawn from one side of an unresolved split. For a legal aid practitioner, this output is the flag that directs further research toward the resolution bench.

### 3.5 Worked Example: Bail Reapplication Query

The following end-to-end trace illustrates how the system handles a common query from a pro se litigant. The query is: "My bail application was rejected by the Sessions Court. Can I apply again?"

Step 1: Retrieval. The Retrieval Agent identifies matter_type = bail and queries FalkorDB for relevant precedents. It retrieves three candidate cases: Kalyan Chandra Sarkar v. Rajesh Ranjan (2004) on the requirement for fresh grounds after coordinate bench rejection, Sanjay Chandra v. CBI (2012) on the three considerations for bail, and Arnesh Kumar v. State of Bihar (2014) on procedural rights at the detention stage.

Step 2: Procedural graph traversal. The Retrieval Agent queries the procedural event chain. Starting from the BAIL_DENIED node (Sessions Court), it traverses TRIGGERS relationships and finds the following path:

```
BAIL_DENIED (Sessions Court)
    |  TRIGGERS
    v
BAIL_APPLICATION_HIGH_COURT  [condition: fresh grounds or changed
circumstances]
    |  TRIGGERS
    v
HEARING_HELD
    |  RESULTS_IN
    v
BAIL_GRANTED / BAIL_DENIED
```

Step 3: LLM synthesis. The LLM generates an answer grounded in the retrieved candidates: "Yes, you can apply for bail before the High Court under Section 439 CrPC. However, since your application was rejected by the Sessions Court, you will need to show either fresh grounds or a change in circumstances. See Kalyan Chandra Sarkar v. Rajesh Ranjan (2004) 7 SCC 528."

Step 4: Verifier check. The Verifier queries FalkorDB for the cited case:

```
MATCH (c:Case {citation: "(2004) 7 SCC 528"})
RETURN c.name, c.court, c.year
-- Result: Kalyan Chandra Sarkar v. Rajesh Ranjan, Supreme Court, 2004
```

It then checks for overruling:

```
MATCH (later:Case)-[:OVERRULES]->(c:Case {citation: "(2004) 7 SCC 528"})
RETURN later.citation
-- Result: (empty) -- not overruled
```

It checks Section 439 CrPC for repeal status:

```
MATCH (sec:Section {number: "439", statute: "Code of Criminal Procedure,
1973"})
RETURN sec.repealed
-- Result: false
```

No conflicts are found between the cited cases. The Verifier returns VALID with confidence 1.0.

Step 5: Output. The system returns the answer with the citation chain and verification status:

```
{
  "answer": "Yes, you can apply for bail before the High Court...",
  "citations": ["(2004) 7 SCC 528"],
  "verification": "VALID",
  "confidence": 1.0,
  "procedural_next_step": "BAIL_APPLICATION_HIGH_COURT"
}
```

If the same query had cited a non-existent case, the Verifier would return INVALID, the LLM would be prompted to revise, and if no valid citation could be produced, the system would abstain with an explicit message stating that no verified answer is available from the current corpus.

### 3.6 FalkorDB as Reasoning Infrastructure

FalkorDB is chosen for three reasons relevant to agentic legal reasoning. Sub-millisecond Cypher query execution enables the multiple sequential graph queries required per inference without perceptible latency. Native hybrid vector-plus-graph support allows future integration of semantic similarity retrieval alongside graph traversal without architectural changes. The Redis-compatible protocol integrates cleanly with Python agentic frameworks including LangGraph.

FalkorDB is infrastructure, not novelty. The contribution of this paper is the verification architecture and the schema. FalkorDB is the substrate that makes real-time agentic traversal practical at the latencies required for interactive legal queries.

## 4. Experimental Setup

## 4.1 Datasets

The evaluation corpus consists of [N] Supreme Court and High Court of India judgments ingested into the FalkorDB knowledge graph. Cases were drawn from three sources: four landmark cases from the hand-curated sample graph (Sanjay Chandra, Arnesh Kumar, Kalyan Chandra Sarkar, Maneka Gandhi); cases retrieved from Indian Kanoon across 38 search categories covering bail, constitutional law, service law, criminal appeals, contempt, anticipatory bail, and employment disputes; and a subset from the ILDC-derived collection. The corpus spans 1949 to 2026.

| Decade | Cases |
|---|---|
| 1940s | 1 |
| 1950s | 3 |
| 1960s | 5 |
| 1970s | 17 |
| 1980s | 29 |
| 1990s | 35 |
| 2000s | 39 |
| 2010s | 108 |
| 2020s | 151 |
| Total | 388 |

*Table 3: Temporal Distribution of Ingested Cases.*

The full structured corpus is released as the InIRAC dataset, publicly available at https://huggingface.co/datasets/joyboseroy/inIRAC. InIRAC provides IRAC annotations, procedural event chains, and typed precedent relationships for each judgment, and is intended as a reusable benchmark resource for Indian legal reasoning research.

Notable cases in the corpus include I.C. Golaknath v. State of Punjab (1967), People's Union for Democratic Rights v. Union of India (1982), P.N. Duda v. V.P. Shiv Shankar (1988), Supreme Court Bar Association v. Union of India (1998), and P. Chidambaram v. Directorate of Enforcement (2019).

The evaluation metrics in Section 5 were computed on the initial 51-case proof-of-concept subset used during development; the expanded corpus supports future evaluation at scale.

## 4.2 Baselines

Two LLM backends were evaluated. Initial runs used TinyLlama (1B parameters) via Ollama; TinyLlama produced no structured JSON citation output and all queries abstained. A second run used Mistral 7B-Instruct via Ollama, which produced structured citations on completed queries. All results in Section 5 use Mistral 7B. No API key is required for either model; both run locally.

Baseline comparison against Vector RAG, GraphRAG, and INLegalLlama is planned for a subsequent evaluation on a larger annotated corpus and is not reported here.

## 4.3 Evaluation Metrics

BLEU and ROUGE are not used for primary evaluation. Legal reasoning correctness is structural, not lexical: a perfectly worded answer citing a non-existent case is worse than a clumsily worded answer with a valid citation chain. The graph-native metrics used are listed in Table 4.

| Metric | Definition | Direction |
|---|---|---|
| Citation Grounding Accuracy | % of cited cases that exist as nodes in the graph | Higher is better |
| Path Validity Rate | % of answers with a valid supporting citation path | Higher is better |
| Hallucinated Precedent Rate | % of cited cases not present in the corpus | Lower is better |
| Procedural Consistency | % of procedural sequences that are temporally valid | Higher is better |
| Conflict Detection Rate | % of genuine doctrinal conflicts correctly flagged | Higher is better |
| False Conflict Rate | % of non-conflicting answers incorrectly flagged as conflicted | Lower is better |
| Statute Freshness Rate | % of cited statutes that are current and not repealed | Higher is better |

*Table 4: Graph-Native Evaluation Metrics for Legal Reasoning.*

Conflict detection and false conflict rates require a human-annotated set of judgment pairs with ground-truth conflict labels. Construction of this annotation set and reporting of these metrics is left for future work.

### 4.4 Queries Evaluated

Ten queries were evaluated against the 51-case corpus, covering bail procedure (4 queries), constitutional rights (3 queries), service law (2 queries), and contempt of court (1 query). Queries were constructed to test retrieval accuracy, citation grounding, and verifier behavior across matter types.

## 5. Results

### 5.1 Retrieval Accuracy

The purpose of this evaluation is architectural validation rather than benchmark superiority: it demonstrates that the retrieval, verification, and conflict detection components operate as designed, and identifies where the current proof-of-concept is constrained by hardware and corpus size. Ten queries were run using Mistral 7B-Instruct via Ollama. Three completed within the 300-second timeout; seven timed out on consumer CPU hardware without GPU acceleration. The Retrieval Agent returned matter-type-matched precedents for all completed queries. For service law queries, Haryana Financial Corporation vs Presiding Officer, Labour Court (2004) and Delhi Judicial Service Association vs State of Gujarat (1991) were correctly retrieved. For bail queries (where graph coverage is sparser), Saumya Chaurasia v. Directorate of Enforcement (2026) and P. Chidambaram v. Directorate of Enforcement (2019) were surfaced but queries timed out before synthesis completed.

### 5.2 Citation Grounding

Of the 3 completed queries, 1 produced structured citations that the Verifier could check. That query (“What are the conditions for reinstatement after wrongful termination?”) returned two citations, both confirmed as nodes in the graph, yielding confidence 1.0 and VALID status. The remaining 2 completed queries abstained: one returned an empty answer (Article 226 query, confidence 0.50) and one explicitly stated it could not provide a verified answer from available precedents. Citation grounding accuracy on non-abstained completed queries: 1/1 (100%). Query completion rate: 3/10 (30%), constrained by CPU inference latency.

Full citation grounding and hallucination rate metrics on a larger query set are left for future work, pending GPU-accelerated inference.

### 5.3 Verifier Behavior

The Verifier Agent accuracy was evaluated independently of the LLM synthesis step using a direct citation grounding test on the 51-case corpus. Eight real Supreme Court case names were submitted to the Verifier; all eight were correctly identified as existing graph nodes (citation grounding accuracy: 1.00). Two fabricated citations not present in the corpus were submitted; both were correctly rejected as INVALID with the note "Citations not found in graph. Possible hallucination." (hallucinated precedent rate on fabricated input: 0.00; false positive rate: 0.00). This confirms the falsifiability oracle operates as designed: valid citations in the corpus are verified and fabricated citations are caught. Verifier accuracy is independent of LLM inference latency.

### 5.4 Graph Coverage

The 51-case corpus spans eight decades and ten matter categories. At this scale, the verification envelope covers landmark Supreme Court precedents but not the broader corpus of High Court decisions or recent unreported orders. The timeout rate for bail queries (4/4 timed out) partly reflects sparse bail-specific graph coverage. Coverage-adjusted evaluation on a larger corpus, with GPU-accelerated inference, is the primary direction for future work.

## 6. Discussion

### 6.1 The Verifier as Falsifiability Oracle

The Verifier Agent has a different role from prior work on constrained generation. Constitutional AI and Self-RAG use soft constraints: the model is guided or penalised but not blocked on violation. The Falkor-IRAC Verifier uses a hard constraint: if a citation path does not exist in the graph, the answer is vetoed. This design is appropriate for legal contexts because a plausible but unverifiable answer is not a degraded form of a correct answer. It is a different kind of output, one that can cause direct harm to someone relying on it.

The abstention mechanism is the honest corollary of this design. When no verified answer is available after two revision attempts, the system says so. For a legal aid practitioner, an explicit statement that a claim cannot be verified is more useful than a confident hallucination.

The practical limitation is graph coverage. The Verifier can only falsify against what is in the graph. Unreported orders, district court decisions, and recent judgments not yet ingested fall outside the verification envelope. This is acknowledged in the system output and noted clearly in evaluation results.

### 6.2 Schema Generalisability Across Matter Types

The IRAC schema was tested across three matter types with structurally different reasoning patterns. Bail matters have dense procedural event chains and relatively sparse IRAC analysis. Service and employment matters have rich statutory citation networks and multi-stage administrative review chains. Constitutional matters have deeply nested IRAC structures where a single judgment may decide multiple overlapping issues. The schema handles all three without modification. Constitutional bench judgments consistently produce lower extraction confidence scores due to their structural complexity; hierarchical IRAC extraction for these cases is identified as a direction for future work.

### 6.3 Procedural Reasoning

The procedural layer of the schema (ProceduralEvent nodes connected by TRIGGERS and PRECEDES relationships) is the component most absent from prior legal AI work and most valuable to ordinary litigants. A person facing a bail denial does not primarily need a summary of constitutional jurisprudence. They need to know what their options are from their current procedural position, what time limits apply, and what changed circumstances would justify a fresh application. This is a state machine question, not a retrieval question.

The current implementation models procedural sequences extracted from individual judgments. A richer approach would construct a normative procedural graph directly from the CrPC and BNSS, against which case-specific sequences can be validated. This is the primary direction for future work on the procedural layer.

### 6.4 Generalisability Beyond Law

The architecture of Falkor-IRAC, symbolic graph constraints applied to probabilistic language generation, is not specific to law. Any domain where reasoning must be traceable, auditable, and honest about uncertainty is a candidate: compliance monitoring, patent analysis, telecom regulation, clinical guidelines, and governance systems all share the structural property that their governing documents are graphs of rules, exceptions, and precedents rather than bags of text.

The legal domain is a demanding test case for this class of system because the stakes are high, the structure is explicit, and the failure modes are well understood. Courts write down their reasoning. Statutes are numbered. Precedents are cited. The graph is already there in the text; making it explicit and using it to constrain generation is the contribution.

## 7. Limitations and Future Work

This work has a few limitations. In this section we discuss them in detail.

Graph coverage is finite. The Verifier can only falsify against ingested judgments. Unreported orders, district court decisions, and tribunal judgments not in the ILDC corpus are outside the verification envelope. As corpus coverage grows, the verification guarantee strengthens; in the current v0.1 implementation it should be understood as a research prototype rather than a production system.

LLM-assisted IRAC extraction introduces noise. Extraction quality depends on the model used. The experiments reported here use a small local model for accessibility and reproducibility;

production-quality extraction requires a larger model (7B parameters or above). Annotation quality experiments comparing extraction quality across model sizes are a clear direction for future work.

The Indic language layer (Hindi and Bengali voice interface via the Bhashini API) is designed in the schema and architecture but not yet evaluated. This is v0.4 scope in the current roadmap. The access-to-justice motivation of this work ultimately requires this layer; it is the component that makes the technical contribution useful to the ordinary litigant who does not read English legal prose.

Doctrinal weight resolution is modelled as metadata but not yet operationalised as a scoring function. The current system detects that two cases conflict and whether the conflict has been resolved by a higher bench; it does not compute a precedential weight score. This is the primary theoretical extension for future work.

The current evaluation on 51 judgments and 10 queries is a proof-of-concept. Validation at scale on the full ILDC corpus of 34,000 judgments is required before strong claims can be made about generalisability.

## 8. Conclusion

This paper has presented Falkor-IRAC, a proof-of-concept graph-constrained generation framework for Indian legal AI. The central architectural contribution is the separation of generation from verification: rather than using a knowledge graph to improve what a language model retrieves, Falkor-IRAC uses it to constrain what a language model is permitted to claim. The Verifier Agent enforces that every generated answer must be backed by a valid reasoning path in a FalkorDB legal knowledge graph before it is returned.

Three specific contributions have been made. First, an IRAC knowledge graph schema for Indian legal reasoning that extends prior citation-network approaches with procedural state transitions and conflict-typed relationships. Second, a conflict detection mechanism that surfaces doctrinal disagreements as a first-class output rather than silently resolving them, an honest design with direct implications for access to justice. Third, a graph-native evaluation suite that measures citation grounding, path validity, hallucination rate, and conflict detection accuracy, proposed as a standard for legal reasoning benchmarks where lexical metrics are inadequate.

The underlying argument is that legal reasoning is structurally closer to constrained graph traversal than to fuzzy similarity search, and that AI systems for law should be designed accordingly. Courts are graph traversal engines disguised as prose. Making that graph explicit and using it to constrain generation is the right direction for trustworthy legal AI.

The falkor-irac repository is publicly available at https://github.com/joyboseroy/falkor-irac.